# Incorporating Expressive Graphical Models in Variational Approximations: Chain-Graphs and Hidden Variables

Tal El-Hay    Nir Friedman
School of Computer Science & Engineering, Hebrew University
{ tale, nir }@cs.huji.ac.il

## Abstract

Global variational approximation methods in graphical models allow efficient approximate inference of complex posterior distributions by using a simpler model. The choice of the approximating model determines a tradeoff between the complexity of the approximation procedure and the quality of the approximation. In this paper, we consider variational approximations based on two classes of models that are richer than standard Bayesian networks, Markov networks or mixture models. As such, these classes allow to find better tradeoffs in the spectrum of approximations. The first class of models are *chain graphs*, which capture distributions that are partially directed. The second class of models are directed graphs (Bayesian networks) with additional latent variables. Both classes allow representation of multi-variable dependencies that cannot be easily represented within a Bayesian network.

## 1 Introduction

A central task in using probabilistic graphical models is *inference*. Exact inference algorithms exploit the structure of the model to decompose the task. In general, although the problem is NP-hard, some structures (e.g., these with bounded tree width) allow efficient inference. When the model is intractable for exact inference, we can still hope to perform approximate inference (although that problem is also known to be generally intractable). One class of approximations that received recent attention is the class of *variational approximation* algorithms [6]. These algorithms attempt to approximate the posterior $P(\mathbf{T} \mid \mathbf{o})$, where $\mathbf{o}$ is an observation of some variable and $\mathbf{T}$ are the remaining variables, by a distribution $Q(\mathbf{T} : \Theta)$ that has tractable structure. Using this approximating distribution, we can define a lower bound on the likelihood $P(\mathbf{o})$. The parameters $\Theta$ that define $Q$ are adapted by trying to maximize this lower bound.

The simplest variational approximation is the *mean-field* approximation [8, 9] that approximates the posterior distribution with a network in which all the random variables are independent. As such, it is unsuitable when there are strong dependencies in the posterior. Saul and Jordan [10] suggest to circumvent this problem by using *structured* variational approximation. This method approximates the posterior by a distribution composed of independent substructures of random variables. This idea can be generalized for various factored forms for $Q$, such as Bayesian networks and Markov networks [1, 11]. Jaakkola and Jordan [5] explore another direction for improving the mean field approximation. They propose to use a mixture of mean field approximations in order to approximate multi-modal posteriors. Both structured variational approximation and the mixture approximation methods allow for a more refined trade-off between accuracy and computational complexity. In the structured approximations more accuracy is gained by adding structure, while in the mixture approximation we can increase the number of mixture components.

In this paper, we generalize and improve on these two methods in order to achieve greater accuracy given the available computational resources. The resulting approximation results enhance the range of approximating distributions and increase the ability to trade-off accuracy for complexity.

We start by considering extensions of structured approximations. Current structured approximation use Bayesian networks *or* Markov networks as approximating distributions. These two classes of models have different expressive power. We provide uniform treatment of both classes by examining *chain graphs* — a class of models that is more expressive than Bayesian and Markov networks, and includes each one of them as a special case.

We then consider how to add *extra* hidden variables to the approximating model. This method generalizes both the structured approximation and the mixture model approximation. It enables us to control the complexity of the approximating model both through the structure and through the number of values of the hidden variables. The extra hidden variables also enable us to maintain the dependency between different variables but control the level of complexity, thus keeping the dependencies in a compressed manner. As straightforward insertion of extra hidden variables to the variational approximation framework results in an intractable optimization problem, we need to combine additional approximation steps. We present a natural generalization of methods suggested by Jaakkola and Jordan [5] for mixtures of mean field models.

## 2 Variational Approximation with Directed Networks

Structured approximations approximate a probability distribution using probability distribution with non-trivial dependency structure. We re-derive standard structured approximation schemes with Bayesian networks (such as the ones in [4, 10, 11]) using tools that will facilitate later developments.

Suppose we are given a distribution $P$ over the set of random variables $\mathbf{X} = \{X_1, \ldots, X_n\}$. Let $\mathbf{O} \subset \mathbf{X}$ be the subset of observed variables. We denote by $\mathbf{T} = \mathbf{X} \setminus \mathbf{O}$ the set of hidden variables. Our task is to approximate the distribution $P(\mathbf{T} \mid \mathbf{o})$ by another distribution $Q(\mathbf{T} : \Theta)$, where $\Theta$ is the set of parameters for the approximating distribution $Q$.

In this paper, we focus on approximating distributions represented by discrete graphical models such as Bayesian networks and Markov networks. That is, we assume that $\{X_1, \ldots, X_n\}$ are discrete random variables and that $P$ has a factorized form

$$P(\mathbf{x}) = \frac{1}{Z_P} \prod_i \phi_i(\mathbf{d}_i) \quad (1)$$

Where $\mathbf{D}_1, \ldots \mathbf{D}_k$ are subsets of $\mathbf{X}$. This representation can be a Bayesian network (in which case, each $\phi_i$ is a conditional distribution) or a Markov network (in which case, each $\phi_i$ is a *potential* over some subset of $\mathbf{X}$).

The approximating distribution will be represented as another graphical model $Q(\mathbf{T} : \Theta)$. Once we specify the form of this model, we wish to find the set of parameters $\Theta$ that minimizes the distance between $Q(\mathbf{T} : \Theta)$ and the posterior distribution $P(\mathbf{T} \mid \mathbf{o})$. A common measure of distance is the *KL divergence* [2] between $Q(\mathbf{T} : \Theta)$ and the posterior distribution $P(\mathbf{T} \mid \mathbf{o})$. This is defined as

$$D(Q(\mathbf{T}) \| P(\mathbf{T} \mid \mathbf{o})) = E_{Q(\mathbf{T})} \left[ \log \frac{Q(\mathbf{T})}{P(\mathbf{T} \mid \mathbf{o})} \right] \quad (2)$$

Finding the parameters for $Q$ will allow us to compute a lower bound for $\log P(\mathbf{o})$. To see this, we define a *functional* $\mathcal{F}$ of the general form:

$$\mathcal{F}[Q \mid \mathbf{c}] = E_{Q(\cdot \mid \mathbf{c})} \left[ \log \frac{P(\mathbf{T}, \mathbf{c}, \mathbf{o})}{Q(\mathbf{T} \mid \mathbf{c})} \right]$$

where $\mathbf{c}$ is an evidence vector assigned to a subset $\mathbf{C} \subseteq \mathbf{T}$, and $Q(\cdot \mid \mathbf{c})$ is a shorthand for $Q(\mathbf{T} \mid \mathbf{c})$. (The reasons for using additional evidence in the definition will be clear shortly.) In the special case where $\mathbf{C} = \emptyset$, $\mathcal{F}$ becomes

$$\mathcal{F}[Q] = E_Q \left[ \log \frac{P(\mathbf{T}, \mathbf{o})}{Q(\mathbf{T})} \right]$$

We can easily verify that

$$\log P(\mathbf{o}) = \mathcal{F}[Q] + D(Q \| P) \geq \mathcal{F}[Q]$$

The inequality is true because the KL divergence is non-negative. Hence, $\mathcal{F}[Q]$ is a lower bound on the log-likelihood. The difference between $\mathcal{F}[Q]$ and the true log-likelihood is the KL-divergence. Minimizing the KL-divergence is equivalent to finding the tightest lower bound.

A simple approximation of this form uses a distribution $Q$ that is a Bayesian network

$$Q(\mathbf{t}) = \prod_j P(x_j \mid \mathbf{u}_j) = \prod_j \theta_{x_j \mid \mathbf{u}_j} \quad (3)$$

where $\mathbf{U}_j$ denotes the parents of $X_j$ in the *approximating* network, and $\theta_{x_j \mid \mathbf{u}_j}$ are the parameters of the distribution.

The computational complexity of calculating the lower bound depends on the complexity of inference in $Q$ and on the domain size of the factors of $P$. To see that let us rewrite $\mathcal{F}[Q]$ in a factorized form.

**Lemma 2.1:** *If* $Q(\mathbf{t}) = \prod_j \theta_{x_j \mid \mathbf{u}_j}$, *then*

$$\mathcal{F}[Q \mid \mathbf{c}] = \sum_i E_{Q(\cdot \mid \mathbf{c})} [\log \phi_i(\mathbf{D}_i, \mathbf{o})] - \log Z_P$$
$$- \sum_j E_{Q(\cdot \mid \mathbf{c})} \left[ \log \theta_{x_j \mid \mathbf{u}_j} \right] + \log Q(\mathbf{c})$$

*where* $\phi_i(\mathbf{D}_i, \mathbf{o})$ *represents a random variable whose value is* $\phi_i(\mathbf{d}_i)$ *if* $\mathbf{d}_i$ *is consistent with* $\mathbf{o}$; *otherwise it is* $0$.

Our goal is to find a set of parameters maximizing $\mathcal{F}$ while conforming to the local normalization constraints. The optimal parameters for $Q$ are found by writing the Lagrangian for this problem and differentiating it with respect to them. The Lagrangian is

$$\mathcal{J}_{BN} = \mathcal{F}[Q] - \sum_j \sum_{\mathbf{u}_j} \lambda_{\mathbf{u}_j} \left( \sum_{x_j} \theta_{x_j \mid \mathbf{u}_j} - 1 \right)$$

To differentiate the Lagrangian we shall use the following technical result.

**Lemma 2.2:** *Let* $Q(\mathbf{t}) = \prod_j \theta_{x_j \mid \mathbf{u}_j}$, *then*

$$\frac{\partial E_Q [f(\mathbf{C})]}{\partial \theta_{x_j \mid \mathbf{u}_j}} = Q(\mathbf{u}_j) \cdot E_{Q(\cdot \mid x_j, \mathbf{u}_j)} [f(\mathbf{C})] + E_Q \left[ \frac{\partial f(\mathbf{C})}{\partial \theta_{x_j \mid \mathbf{u}_j}} \right]$$

**Corollary 2.3:**

$$\frac{\partial \mathcal{F}[Q]}{\partial \theta_{x_j \mid \mathbf{u}_j}} = Q(\mathbf{u}_j) \cdot (\mathcal{F}[Q \mid x_j, \mathbf{u}_j] - \log Q(x_j, \mathbf{u}_j) - 1)$$

Note that $\log Q(x_j, \mathbf{u}_j) = \log Q(\mathbf{u}_j) + \log \theta_{x_j \mid \mathbf{u}_j}$. Equating the derivative of the Lagrangian to zero and dividing both sides by $Q(\mathbf{u}_j)$ and then rearranging, we get

$$\theta_{x_j \mid \mathbf{u}_j} = \frac{1}{Z_{\mathbf{u}_j}} \cdot e^{\mathcal{E}_{BN}(x_j, \mathbf{u}_j)} \quad (4)$$

where $Z_{\mathbf{u}_j}$ is a normalization constant, and

$$\mathcal{E}_{BN}(x_j, \mathbf{u}_j) = \mathcal{F}[Q \mid x_j, \mathbf{u}_j] - \log Q(\mathbf{u}_j) \quad (5)$$
$$= E_{Q(\cdot \mid x_j, \mathbf{u}_j)} \left[ \sum_i \log \phi_i(\mathbf{D}_i, \mathbf{o}) - \sum_{j' \neq j} \log \theta_{x_{j'} \mid \mathbf{u}_{j'}} \right]$$
$$- \log Z_P$$

To better understand this characterization, we examine the term $\mathcal{F}[Q \mid x_j, \mathbf{u}_j]$. It is easy to verify that

$$D(Q(\mathbf{T} \mid x_j, \mathbf{u}_j) \| P(\mathbf{T} \mid x_j, \mathbf{u}_j, \mathbf{o})) =$$
$$-\mathcal{F}[Q \mid x_j, \mathbf{u}_j] + \log P(x_j, \mathbf{u}_j, \mathbf{o})$$

Thus, $\mathcal{F}[Q \mid x_j, \mathbf{u}_j]$ is a lower bound on $\log P(x_j, \mathbf{u}_j, \mathbf{o})$. This suggests that Eq. 4 can be thought of as approximating $P(x_j \mid \mathbf{u}_j, \mathbf{o})$ by $\theta_{x_j \mid \mathbf{u}_j}$. If we replace $\mathcal{E}_{BN}(x_j, \mathbf{u}_j)$ by $\log P(x_j, \mathbf{u}_j, \mathbf{o})$ in this equation, we would get that $\theta_{x_i \mid \mathbf{u}_i} = P(x_j \mid \mathbf{u}_j, \mathbf{o})$.[1]

In order to find optimal parameters, we can use an iterative procedure that updates the parameters of one family on each iteration. An asynchronous update of the parameters according to Eq. 4 guarantees a monotonic increase in the lower bound $\mathcal{F}[Q]$ and converges to a local maximum. This is a consequence of the fact that, for every $i$ and every assignment to the parents $\mathbf{u}_j$, $\mathcal{F}$ is a concave function of the set of parameters $\{\theta_{x_j \mid \mathbf{u}_j} \mid x_j \in dom(X_j)\}$. Therefore, the stationary point is a global maximum with respect to those parameters. The concavity of $\mathcal{F}$ follows from the fact that the second order partial derivatives are negative

$$\frac{\partial^2 \mathcal{F}}{\partial \theta_{x_j \mid \mathbf{u}_j}^2} = -\frac{1}{\theta_{x_j \mid \mathbf{u}_j}} Q(\mathbf{u}_j) < 0$$

and the mixed partial derivatives are all zero.

The complexity of calculating $\mathcal{E}_{BN}$ as defined in Eq. 5 is determined by the number of variables, the size of the families in $P$ and by the complexity of calculating marginal probabilities in $Q$. It is important to realize that not all the terms in this equation need to be computed. To see this we need to consider *conditional independence* properties in $Q$. We say that $X$ is independent of $\mathbf{Y}$ given $\mathbf{Z}$ in $Q$, denoted $Q \models Ind(X; \mathbf{Y} \mid \mathbf{Z})$, if $Q(X \mid \mathbf{y}, \mathbf{z}) = Q(X \mid \mathbf{z})$ for all values $\mathbf{y}$ and $\mathbf{z}$ of $\mathbf{Y}$ and $\mathbf{Z}$. If $Q$ is a Bayesian network, we can determine such independencies using *d-separation* [7]. Now, suppose that $Q \models Ind(X_j; \mathbf{C} \mid \mathbf{U}_j)$, i.e., $Q(\mathbf{c} \mid x_j, \mathbf{u}_j) = Q(\mathbf{c} \mid \mathbf{u}_j)$. Terms of the form $E_{Q(\mathbf{c} \mid x_j, \mathbf{u}_j)}[f(\mathbf{c})]$ can be ignored in the update equations since they change the new parameters by a constant factor which will be absorbed in $Z_{\mathbf{u}_j}$. Therefore we can reduce the amount of computations by defining the sets of indices of the factors that depend on $X_j$ given $\mathbf{U}_j$ as follows:

$$F_j^p = \{i : Q \not\models Ind(X_j; \mathbf{D}_i \mid \mathbf{U}_j)\}$$
$$F_j^q = \{j' \neq j : Q \not\models Ind(X_j; X_{j'}, \mathbf{U}_{j'} \mid \mathbf{U}_j)\}$$

[1] Note that the term $\log Q(\mathbf{u}_j)$ in $\mathcal{E}_{BN}(x_j, \mathbf{u}_j)$ can be ignored, since it is absorbed by the normalizing constant. We include it above to simplify the decomposition of $\mathcal{E}_{BN}(x_j, \mathbf{u}_j)$.

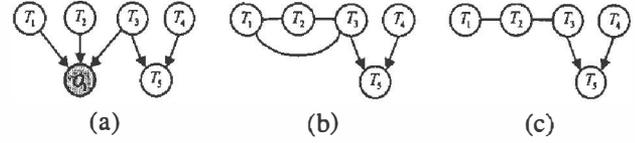

(a) (b) (c)

Figure 1: (a) A Bayesian network with an observed variable ($O_1$). (b) A representation of the posterior distribution as a chain graph. (c) an approximating chain graph network.

We can then redefine $\mathcal{E}_{BN}$ to be

$$\mathcal{E}_{BN}(x_j, \mathbf{u}_j) = \sum_{i \in F_j^p} E_{Q(\cdot \mid x_j, \mathbf{u}_j)} [\log \phi_i(\mathbf{D}_i, \mathbf{o})] -$$
$$\sum_{j' \in F_j^q} E_{Q(\cdot \mid x_j, \mathbf{u}_j)} [\log Q(X_{j'} \mid \mathbf{U}_{j'})]$$

Depending on the decomposition of $Q$, this formula might involve much fewer terms then Eq. 5. For example, in mean field approximation, $F_j^p$ includes only potentials that include $X_j$, and $F_j^q$ is empty.

Similar derivation can be made when $Q$ is a Markov network. The main difference is that in Markov networks there is a global constraint (defined by the partition function) rather then local ones for each conditional distribution. Due to space considerations we omit the details, and refer the interested reader to [11].

## 3 Chain Graph Approximations

As is well known, the classes of distributions that can be represented by Markov networks and by Bayesian networks are not equivalent. Therefore, for some distributions the best tractable approximations might be represented by Bayesian networks while for other distributions the best approximation is a Markov network. We can gain more flexibility in choosing an approximating distribution by using a more general class of probability models that can capture the dependency models implied by Bayesian networks, Markov networks and dependency models that can be captured by neither of them.

To consider a concrete example, suppose that $P$ is a Bayesian network. What is the form of the posterior $P(\mathbf{T} \mid \mathbf{o})$? For a concrete example, consider the network of Figure 1(a). When, we observe the value of $O_1$, we create dependencies among the variables $T_1, T_2$, and $T_3$. The posterior distribution is neither a Bayesian network nor a Markov network (because of the v-structure in the parents of $T_5$). Instead, we can write this posterior in the form:

$$\psi(T_1, T_2, T_3) P(T_1) P(T_2) p(T_3) p(T_4) P(T_5 \mid T_3, T_4)$$

where $\psi(T_1, T_2, T_3) = \frac{1}{P(o_1)} P(o_1 \mid T_1, T_2, T_3)$ is a potential that is induced by the observation of $o_1$.

A natural class of models that has this general form are *chain graphs* [3]. Such a model factorizes to a product of conditional distributions and potentials. Formally, we define a chain graph to have for each variable a (possibly empty) set of parents, and in addition to have a set of potentials on some subsets of variables.

When we represent $Q$ as a chain graph, we will have the general form:

$$Q(\mathbf{T}) = \frac{1}{Z_Q} \prod_j Q(x_j \mid \mathbf{u}_j) \prod_k \psi_k(C_k)$$

where, as before, $\mathbf{U}_j$ are the directed parents of $X_j$. In addition, $\psi_k$ are potential functions on subsets of $\mathbf{T}$, and $Z_Q = \sum_\mathbf{t} \prod_j Q(x_j \mid \mathbf{u}_j) \prod_k \psi_k(C_k)$ is a normalizing function that ensures that the distribution sums to 1. Figure 1(b) shows the chain graph that represents this factorization.

It is easy to check that if $P$ is a Bayesian network, then $P(\mathbf{T} \mid \mathbf{o})$ can be represented as a chain graph (for each variable $X_j$ in $\mathbf{O}$, add a potential over the parents of $X_j$). In contrast, it is easy to build examples where the posterior distribution cannot be represented by a Bayesian network without introducing unnecessary dependencies. Thus, this class of models is, in some sense, a natural representation of conditional distributions in Bayesian networks.

This argument suggests that by considering chain graphs we can represent approximate distributions that are more tractable than the original distribution, yet are closer to the posterior we want to approximate. For example, Figure 1(c) shows a simple example for a possible approximate network for representing the posterior of the network of Figure 1(a). In this network there are two potentials with two variables each, rather than one with three variables.

Given the structure of the approximating chain graph, we wish to find the set of parameters that maximizes $\mathcal{F}[Q]$, the lower bound on the log-likelihood. As usual, we need to define a Lagrangian that capture the constraints on the model. These constraints contain the constraints that appeared in the Bayesian network case, and, in addition, we require that each potential sums up to one:

$$\sum_{c_k} \psi_k(c_k) = 1 \quad \forall k$$

To understand this constraint, note that the each potential can be scaled without changing $Q$, since the scaling constant is absorbed in $Z_Q$. Thus, without constraining the scale of each potential there is a continuum of solutions, and the magnitude of values in the potentials can explode.

Putting these together, the Lagrangian has the form:

$$\mathcal{J}_{CG} = \mathcal{F}[Q] - \sum_j \sum_{\mathbf{u}_i} \lambda_{\mathbf{u}_j} \sum_{x_j} \theta_{x_j \mid \mathbf{u}_j} - \sum_k \lambda_k \sum_{c_k} \psi_k(c_k)$$

The main difference from the Bayesian network approximation is in the form of the analogue of Lemma 2.2. In the case of chain graphs, we also have to differentiate $Z_Q$, and so we get slightly more complex derivatives.

**Lemma 3.1:** *If $Q$ is a chain graph over $\mathbf{T}$, then*

$$\frac{\partial E_Q[f(\mathbf{C})]}{\partial \theta_{x_j \mid \mathbf{u}_j}} = E_Q\left[\frac{\partial f(\mathbf{C})}{\partial \theta_{x_j \mid \mathbf{u}_j}}\right]$$
$$+ \frac{Q(x_j, \mathbf{u}_j)}{\theta_{x_j \mid \mathbf{u}_j}} \cdot \left(E_{Q(\cdot \mid x_j, \mathbf{u}_j)}[f(\mathbf{C})] - E_Q[f(\mathbf{C})]\right)$$
$$\frac{\partial E_Q[f(\mathbf{C})]}{\partial \psi_k(c_k)} = E_Q\left[\frac{\partial f(\mathbf{C})}{\partial \psi_k(c_k)}\right]$$
$$+ \frac{Q(\mathbf{c}_k)}{\psi_k(c_k)} \cdot \left(E_{Q(\cdot \mid \mathbf{c}_k)}[f(\mathbf{C})] - E_Q[f(\mathbf{C})]\right)$$

Note that when we differentiate $\mathcal{F}[Q]$ we get two terms. The first, is $\mathcal{F}[Q \mid x_j, \mathbf{u}_j]$ as before, and the other is $\mathcal{F}[Q]$. However, since $\mathcal{F}[Q]$ does not depend on the value of $x_j$, it is a absorbed in the normalizing constant $Z_{\mathbf{u}_j}$. Thus, the general structure of the solution remains similar to the simpler case of Bayesian networks:

$$\theta_{x_j \mid \mathbf{u}_j} = \frac{1}{Z_{\mathbf{u}_j}} e^{\mathcal{E}_{CG}(x_j, \mathbf{u}_j)}$$
$$\psi_k(c_k) = \frac{1}{Z_{\mathbf{c}_k}} e^{\mathcal{E}_{CG}(\mathbf{c}_k)}$$

where

$$\mathcal{E}_{CG}(x_j, \mathbf{u}_j) = \mathcal{F}[Q \mid x_j, \mathbf{u}_j] - \log Q(x_j, \mathbf{u}_j) + \log \theta_{x_j \mid \mathbf{u}_j}$$
$$\mathcal{E}_{CG}(\mathbf{c}_k) = \mathcal{F}[Q \mid \mathbf{c}_k] - \log Q(\mathbf{c}_k) + \log \psi_k(c_k)$$

To get an explicit form of these equations, we simply write the chain-graph analogue of Lemma 2.1 which has similar form but includes additional terms. As in the case of Bayesian network, we can easily identify terms that can depend on the value of $x_j$, and focus the computation only on these. This is a straightforward extension of the ideas in Bayesian networks, and so we omit the details.

## 4 Adding Hidden Variables

Structured approximations were the first method proposed for improving the mean field approximation. Jaakkola and Jordan [5] proposed another way of improving the mean field approximation: to use mixture distributions, where each mixture component is represented by a factorized distribution. The motivation for using mixture distribution emerges from the fact that in many cases the posterior distribution is multi-modal, i.e. there are several distinct regions in the domain of the distribution with relatively high probability values. If the location of the different modes of the distribution depends on the values of several variables than the mean field approximation can not capture more than one mode.

Recall that the mean field approximation uses a graphical model in which all the variables in $\mathbf{T}$ are independent of each other. Thus, we can think of it as a Bayesian network without edges. The mixture distribution approximation can be viewed as one that uses a Bayesian network over the variables $\mathbf{T}$ and an extra variable $V$, such that $V$ is the parent node of each $X_j \in \mathbf{T}$. As before, the parameters of the mixture distribution could be found by maximizing the lower bound of the log likelihood as presented in Section 2. Unfortunately, using this technique in a straightforward manner would not help us since the extra hidden variables introduces correlations, which leave us with an optimization problem whose complexity is at least as great as this of the original inference problem. Jaakkola and Jordan overcame this problem by introducing another variational transformation resulting in another lower bound to the log likelihood [5].

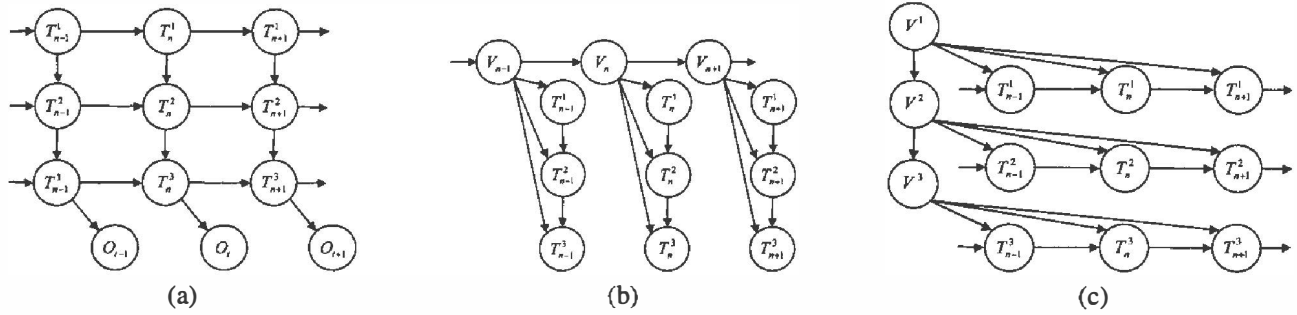

Figure 2: (a) A simple dynamic Bayesian network that describe a temporal process. Time progresses to the right. Each vertical "slice" describe variables that exist in the same instance. (b) and (c) are two approximating networks for the distribution represented by the network (a) with extra hidden variables. (b) Edges within a time slice are maintained. Correlations between time slice are modeled through the introduction of the hidden variable set $\{V_n\}_{n=1}^{N}$. (c) Edges between time slices are maintained. Correlations between the three chains are modeled through the hidden variables $V^1$, $V^2$ and $V^3$.

In this section, we generalize the ideas of Jaakkola and Jordan, and show a method where we can perform structured approximation with distributions $Q$ that are defined over $\mathbf{T} \cup \mathbf{V}$, where $\mathbf{V}$ is a set of hidden variables that did not appear in the original distribution. (For clarity, we focus on the case of Bayesian networks, although similar extension can be applied to chain graphs as well.)

Given the distribution $P(\mathbf{X})$ and evidence o we shall approximate the posterior $P(\mathbf{T} \mid \mathbf{o})$ with another distribution $Q(\mathbf{T}) = \sum_{\mathbf{v}} Q(\mathbf{T}, \mathbf{v})$. This distribution is defined over the variable set $\mathbf{T} \cup \mathbf{V}$ where $\mathbf{V}$ is a set of extra hidden variables. Our task is to find the parameters of $Q$ that will maximize the lower bound $\mathcal{F}[Q]$.

Figure 2(b) and (c) are two examples of possible approximations for the distribution that is represented by the network in Figure 2(a). Recall the structured approximation for this network modeled the approximating distribution by a network with three independent chains. In the networks presented here, the correlations are maintained through the hidden variables. In Figure 2(a) we added an extra hidden variable for every time slice. The correlations between time slices are maintained through those hidden variables. The edges within a time slice are maintained in order to preserve intra-time dependencies. In Figure 2(b) we maintained the edges between the time slices and added extra hidden variables for every chain. Correlations among the chains are maintained by the connections between the hidden variables.

Another perspective of the potential of extra hidden variables was suggested by Jaakkola and Jordan [5]. We can easily extend it to our case. This is done by reexamination of the lower bound $\mathcal{F}[Q]$.

**Lemma 4.1:** Let $Q(\mathbf{T}) = \sum_{\mathbf{v}} Q(\mathbf{T}, \mathbf{v})$, then

$$\mathcal{F}[Q] = E_{Q(\mathbf{V})}\left[\mathcal{F}[Q \mid \mathbf{V}]\right] + I(\mathbf{T}; \mathbf{V})$$

where $I(\mathbf{T}; \mathbf{V}) = E_Q\left[\log \frac{Q(\mathbf{T}|\mathbf{V})}{Q(\mathbf{T})}\right]$ is the mutual information between $\mathbf{T}$ and $\mathbf{V}$.

The first term is an average on lower bounds that are gained without introducing extra hidden variables. The improvement arises from the second term. Given the structure of the approximating network without extra hidden variables, the lower bound can be improved if there are several different configurations of the parameters of the sub-network defined on $\mathbf{T}$ that achieve lower bounds that are near optimal. Using an extra hidden variable set to combine these configuration, will improve the lower bound by the amount of the mutual information between $\mathbf{T}$ and $\mathbf{V}$.

As described above, in the presence of hidden variable, the optimization of the functional $\mathcal{F}[Q]$ is more complex. The source of these complications is the fact that $\log Q(\mathbf{t})$ does not decompose. Therefore we shall relax the lower bound. We start by rewriting $\mathcal{F}[Q]$ as

$$\mathcal{F}[Q] = E_Q\left[\log \frac{P(\mathbf{T}, \mathbf{o})}{Q(\mathbf{T}, \mathbf{V})}\right] - H(\mathbf{V} \mid \mathbf{T})$$

This first term does decompose. The remaining term is the *conditional entropy*

$$H(\mathbf{V} \mid \mathbf{T}) = E_Q\left[\log \frac{Q(\mathbf{T})}{Q(\mathbf{T}, \mathbf{V})}\right]$$

Instead of decomposing this term, we can calculate a lower bound for it by introducing extra variational parameters. The new parameters are based on the convexity bound [6]

$$-\log(x) \geq -\lambda x + \log(\lambda) + 1 \quad (6)$$

We can use the convexity bound by adding an extra variational parameter $R(\mathbf{t}, \mathbf{v})$ for every assignment to $\mathbf{T} \cup \mathbf{V}$. Applying Equation 6 for every term in the summation of the conditional entropy, we get a lower bound for the conditional entropy:

$$-H(\mathbf{V} \mid \mathbf{T})$$
$$\geq E_Q\left[-R(\mathbf{t},\mathbf{v})\frac{Q(\mathbf{t})}{Q(\mathbf{t},\mathbf{v})} + \log R(\mathbf{t},\mathbf{v}) + 1\right]$$
$$= -\sum_{\mathbf{t},\mathbf{v}} R(\mathbf{t},\mathbf{v})Q(\mathbf{t}) + E_Q[\log R(\mathbf{t},\mathbf{v})] + 1$$

Obviously, if we add a distinct variational parameter for every assignment $\mathbf{t}, \mathbf{v}$, the conditional entropy can be recovered accurately. Unfortunately, this setting leaves us with an intractable computation. In order to reduce the computational complexity of the lower bound, we assume that $R$ has a similar structure to that of $Q$

$$R(\mathbf{t}, \mathbf{v}) = \prod_j \rho_{x_j, \mathbf{u}_j}$$

We define the lower bound on $\mathcal{F}$ as a new functional:

$$\mathcal{G}[Q, R \mid \mathbf{c}] = E_{Q(\cdot \mid \mathbf{c})}\left[\log \frac{P(\mathbf{T}, \mathbf{c}, \mathbf{o})}{Q(\mathbf{T}, \mathbf{V} \mid \mathbf{c})}\right]$$
$$- \sum_{\mathbf{v}} E_{Q(\mathbf{T} \mid \mathbf{c})}[R(\mathbf{T}, \mathbf{v})]$$
$$+ E_{Q(\cdot \mid \mathbf{c})}[\log R(\mathbf{T}, \mathbf{V})] + 1$$

We now can define the Lagrangian with the desired constraints:

$$\mathcal{J}_H = \mathcal{G}[Q, R] - \sum_j \sum_{\mathbf{u}_j} \lambda_{\mathbf{u}_j} \sum_{x_j} \theta_{x_j \mid \mathbf{u}_j}$$

Using Lemma 2.2, and then applying the constraints, we get the update equations for $\theta_{x_j \mid \mathbf{u}_j}$:

$$\theta_{x_j \mid \mathbf{u}_j} = \frac{1}{Z_{\mathbf{u}_j}} \cdot e^{\mathcal{E}_H(x_j, \mathbf{u}_j)} \quad (7)$$

Where

$$\mathcal{E}_H(x_j, \mathbf{u}_j) = \mathcal{G}[Q, R \mid x_j, \mathbf{u}_j] - \log(\mathbf{u}_j)$$

As usual, we can decompose this term to a sum of terms:

$$\mathcal{E}(x_j, \mathbf{u}_j) = \sum_i E_{Q(\cdot \mid x_j, \mathbf{u}_j)}[\log \phi_i(\mathbf{D}_i, \mathbf{o})]$$
$$- \sum_{j' \neq j} E_{Q(\cdot \mid x_j, \mathbf{u}_j)}[\log Q(X_{j'} \mid \mathbf{U}_{j'})]$$
$$+ \sum_{j'} E_{Q(\cdot \mid x_j, \mathbf{u}_j)}[\log R(X_{j'} \mid \mathbf{U}_{j'})]$$
$$- \sum_{\mathbf{v}} E_{Q(\mathbf{T} \mid x_j, \mathbf{u}_j)}[R(\mathbf{T}, \mathbf{v})]$$

the expression $\mathcal{E}_H$ is similar to the one obtained for the simpler structural approximation, except for the last two terms that arise from the lower bound on the negative conditional entropy. To evaluate the term $\sum_{\mathbf{v}} E_{Q(\mathbf{T} \mid x_j, \mathbf{u}_j)}[R(\mathbf{T}, \mathbf{v})]$ we perform variable-elimination like dynamic programming algorithm.

To complete the story, we need to consider the update equations for the parameters of $R$. Simple differentiation results in the equation

$$\rho_{x_i, \mathbf{u}_i} = \frac{\rho_{x_i, \mathbf{u}_i}}{\sum_{\mathbf{t}, \mathbf{v} \models x_j, \mathbf{u}_j} R(\mathbf{t}, \mathbf{v}) Q(\mathbf{t})} Q(x_j, \mathbf{u}_j) \quad (8)$$

Where the term $\mathbf{t}, \mathbf{v} \models x_j, \mathbf{u}_j$ denotes assignments to $\{\mathbf{T}, \mathbf{V}\}$ that are consistent with $x_j, \mathbf{u}_j$. Note that $\rho_{x_i, \mathbf{u}_i}$ does not appear in the right hand side (since it cancels out in the fraction). Again, we can efficiently compute such equations using dynamic programming.

The Lagrangian is a convex function of both $\theta_{x_j \mid \mathbf{u}_j}$ and $\rho_{x_j, \mathbf{u}_j}$. Therefore, asynchronous iterations of Equation 8 and Equation 7 improve the lower bound and will eventually converge to a stationary point.

## 5 Examples

To evaluate our methods we performed a preliminary test with synthetic data. We created dynamic Bayesian networks with the general architecture shown in Figure 2(a). All the variables in these networks are binary. We controlled two parameters: the number of time slices expanded, and the number of variables in each slice. The parameters of networks were sampled from a Dirichlet prior with hyperparameter $\frac{1}{2}$. Thus, there was some bias toward skewed distributions. Our aim was to compute the likelihood of the observation in which all observed variables were set to be 0. We repeated these tests for sets of 20 networks sampled for each combination of the two parameters (number of time slices and number of variables per slice).

We performed variational approximation to the posterior distribution using three types of networks with hidden variables: The first two types are based on the "vertical" and "horizontal" architectures shown in Figure 2(b) and (c). We considered networks with 1, 2, and 3 values for the hidden variable. (Note that when we consider a hidden variable with one value, we essentially apply the Bayesian network structured approximation.) The third type are networks that represent mixture of mean field approximations. For this type we considered networks with 1, 4 and 6 mixture components (When there is one mixture component the approximation is simply mean field). We run each procedure for 10 iterations of asynchronous updates. This seems to converge on most runs. To avoid local maxima, we tried 10 different random starting points in each run and returned the best scoring one.

The figure of merit for our approximations is the reported upper-bound on the KL-divergence between the approximation and the true posterior. This is simply $\log P(\mathbf{o})$ − $\mathcal{G}_Q[Q, R]$. (The examples are sufficiently small, so that we can compute $\log P(\mathbf{o})$.) We need to examine this quantity since different random networks have different values of $P(\mathbf{o})$ and so we cannot compare lower bounds.

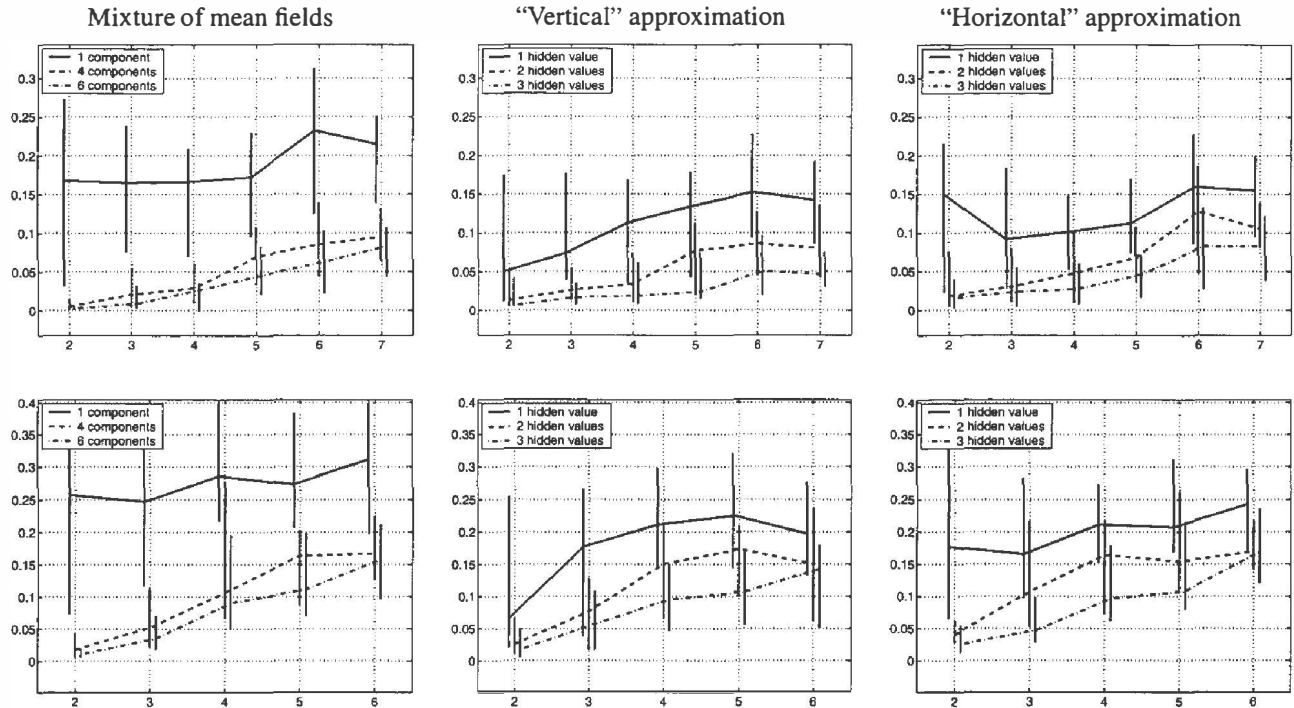

Figure 3: Comparison of the two approximating structures of Figure 2 and mixture of mean fields. The figures on the left column report results for the mixture of mean fields approximation, with 1, 4, and 6 mixture components. The figures on the middle column report results for the network structure containing additional hidden variable for each time slice (Figure 2(b)) with hidden variables with 1, 2, and 3 values. The figures on the right column report results for the network structure containing additional hidden variable for each temporal chain (Figure 2(c)) with hidden variables with 1, 2, and 3 values. The figures on the top row report on approximation to networks with 3 variables per time slice and the figures on the bottom row report on networks with 4 variables per time slice. The $x$-axis corresponds to the number of time slices in the network. The $y$-axis corresponds to the upper-bound on the KL-divergence $\log P(\mathbf{o}) - \mathcal{G}_Q[Q, R]$ normalized by the number of time slices in the network. Lines describe to median performance among 20 inference problems, and error bars describe 25-75 percentiles.

Figure 3 describes the results of these runs. As we can see the differences grow with the number of time slices. This is expected as the problem becomes harder with additional slices. The general trend we see is that runs with more hidden values perform better. These differences are mostly pronounced in the larger networks. This is probably due to the higher complexity of these networks.

The comparison to mixture of mean fields approximation shows that simple mean field (1 component) is much worse than all the other methods. Second, we see that although mixtures of mean field improve with larger number of components, they are still worse than the structured approximations on the network with 3 variables per slices. We believe that these toy examples are not sufficiently large to highlight the differences between the different methods. For example, differences start to emerge when we examine 6 and 7 time slices.

Our implementation of these variational methods is not optimized and thus we do not believe that running times are informative on these small examples. Nonetheless, we note that running mixtures of mean fields with 6 components took roughly the same time as running the structured approximations with hidden variables of cardinality 3.

One caveat of this experiment is that it is based on random networks, for which the depenencies between variables is often quite weak. As such it is hard to gauge how hard is inference in this networks. We are currently starting to apply these methods to real-life problems, where we expect to improvement over mean field type methods to be more pronounced.

## 6 Discussion

In this paper we presented two extensions of structured variational methods—based on chain graphs and additional hidden variables. Each extension exploits a representa-tional feature that allows to better match a tractable approximating network to the posterior. By perusing such extensions we hope to find better tradeoffs between network complexity on one hand and the approximation of the posterior distribution on the other.

We demonstrated the effect of using hidden variables in synthetic examples and showed that learning non-trivial network with hidden variables helps the approximation. We are currently starting larger scale experiments on hard real-life problems.

We put emphasis on presenting uniform machinery in the derivations of the three variants we considered. This uniform presentation allows for better insights into the workings of such approximations and simplifies the process of deriving new variants for other representations.

One issue that we did not address here for lack of space is efficient computations on the network $Q$. The usual analysis focuses on the maximal tree width of the network. However, much computation (up to a quadratic factor) can be saved by conscious planning of order of asynchronous updates and the propagation of messages in $Q$'s join tree.

The grand challenge for applications of such variational methods is to build automatic procedures that can determine what structure matches best a given network with a given query. This is a non-trivial problem. We hope that some of the insights we got from our derivations can provide initial clues that will lead to development of such methods.

### Acknowledgements

We thank Gal Elidan, Tommy Kaplan, Iftach Nachman, Matan Ninio, and the anonymous referees for comments on earlier drafts of this paper. This work was supported in part by Israeli Science Foundation grant number 224/99-1. Nir Friedman was also supported by an Alon fellowship and by the Abe & Harry Sherman Senior Lectureship in Computer Science. Experiments reported here were run on equipment funded by an ISF Basic Equipment Grant.